\documentclass{article}

    \PassOptionsToPackage{round}{natbib}
\usepackage[final]{neurips_glfrontiers_2023}

\usepackage[utf8]{inputenc} 
\usepackage[T1]{fontenc}    
\usepackage{hyperref}       
\usepackage{url}            
\usepackage{booktabs}       
\usepackage{amsfonts}       
\usepackage{nicefrac}       
\usepackage{microtype}      
\usepackage[x11names]{xcolor}         
\usepackage{bm, mathtools, amsmath, amssymb}
\usepackage[capitalize, noabbrev, nameinlink]{cleveref}
\usepackage{array,multirow}
\usepackage{wrapfig}
\usepackage{graphicx}
\usepackage{tikz}
\usepackage{pgfplots}
\usepackage[font=small,labelfont=bf]{caption}
\pgfplotsset{compat=1.17}

\hypersetup{
  breaklinks,
  colorlinks,
  linkcolor=IndianRed3,
  citecolor=DodgerBlue4,
  urlcolor=DodgerBlue4,
}

\newcommand\pcref[1]{(\cref{#1})}

\definecolor{first}{HTML}{FF0000}
\definecolor{second}{HTML}{FF7B00}
\definecolor{third}{HTML}{A556EB}

\newcommand\first[1]{\textbf{#1}}
\newcommand\second[1]{\underline{#1}}
\newcommand\third[1]{\textit{#1}}

\newcommand\nnfootnote[2]{%
  \begin{NoHyper}
  \renewcommand\thefootnote{#1}\footnotetext{#2}%
  \addtocounter{footnote}{-1}%
  \end{NoHyper}
}

\def\relDrop{{\gR_{\text{adv}}^{\text{clean}}}}

\def\eqref#1{equation~\ref{#1}}

\def\1{\bm{1}}

\def\defeq{{\vcentcolon=}}

\def\vh{{\bm{h}}}

\def\vm{{\bm{m}}}

\def\vs{{\bm{s}}}

\def\vx{{\bm{x}}}
\def\vy{{\bm{y}}}

\def\evm{{m}}

\def\mA{{\bm{A}}}

\def\mH{{\bm{H}}}

\def\mK{{\bm{K}}}

\def\mX{{\bm{X}}}

\DeclareMathAlphabet{\mathsfit}{\encodingdefault}{\sfdefault}{m}{sl}
\SetMathAlphabet{\mathsfit}{bold}{\encodingdefault}{\sfdefault}{bx}{n}

\def\gB{{\mathcal{B}}}
\def\gC{{\mathcal{C}}}
\def\gD{{\mathcal{D}}}
\def\gE{{\mathcal{E}}}

\def\gG{{\mathcal{G}}}
\def\gH{{\mathcal{H}}}
\def\gI{{\mathcal{I}}}

\def\gL{{\mathcal{L}}}

\def\gP{{\mathcal{P}}}

\def\gR{{\mathcal{R}}}

\def\gT{{\mathcal{T}}}

\def\gY{{\mathcal{Y}}}

\def\emA{{A}}

\newcommand{\E}{\mathbb{E}}

\newcommand{\R}{\mathbb{R}}

\DeclareMathOperator*{\argmax}{arg\,max}

\title{On the Adversarial Robustness of Graph Contrastive Learning Methods}

\author{%
  Filippo Guerranti, Zinuo Yi$^*$, Anna Starovoit$^*$, Rafiq Kamel$^*$,\\\textbf{Simon Geisler, Stephan Günnemann} \\
  \texttt{\{f.guerranti, z.yi, r.kamel, a.starovoit, s.geisler, s.guennemann\}@tum.de} \\
  Department of Computer Science \& Munich Data Science Institute \\ Technical University of Munich \\
}

\begin{document}
    \maketitle
    \nnfootnote{$*$}{Equal contribution.}
    \begin{abstract}
Contrastive learning (CL) has emerged as a powerful framework for learning representations of images and text in a self-supervised manner while enhancing model robustness against adversarial attacks. More recently, researchers have extended the principles of contrastive learning to graph-structured data, giving birth to the field of graph contrastive learning (GCL). However, whether GCL methods can deliver the same advantages in adversarial robustness as their counterparts in the image and text domains remains an open question.
In this paper, we introduce a comprehensive \textit{robustness evaluation protocol} tailored to assess the robustness of GCL models. We subject these models to \textit{adaptive} adversarial attacks targeting the graph structure, specifically in the evasion scenario. We evaluate node and graph classification tasks using diverse real-world datasets and attack strategies. With our work, we aim to offer insights into the robustness of GCL methods and hope to open avenues for potential future research directions.
\end{abstract}

    \section{Introduction}

 \begin{wrapfigure}{11}{0.37\textwidth}
    \centering
    \vspace{-1.2em}
    \includegraphics[width=\linewidth]{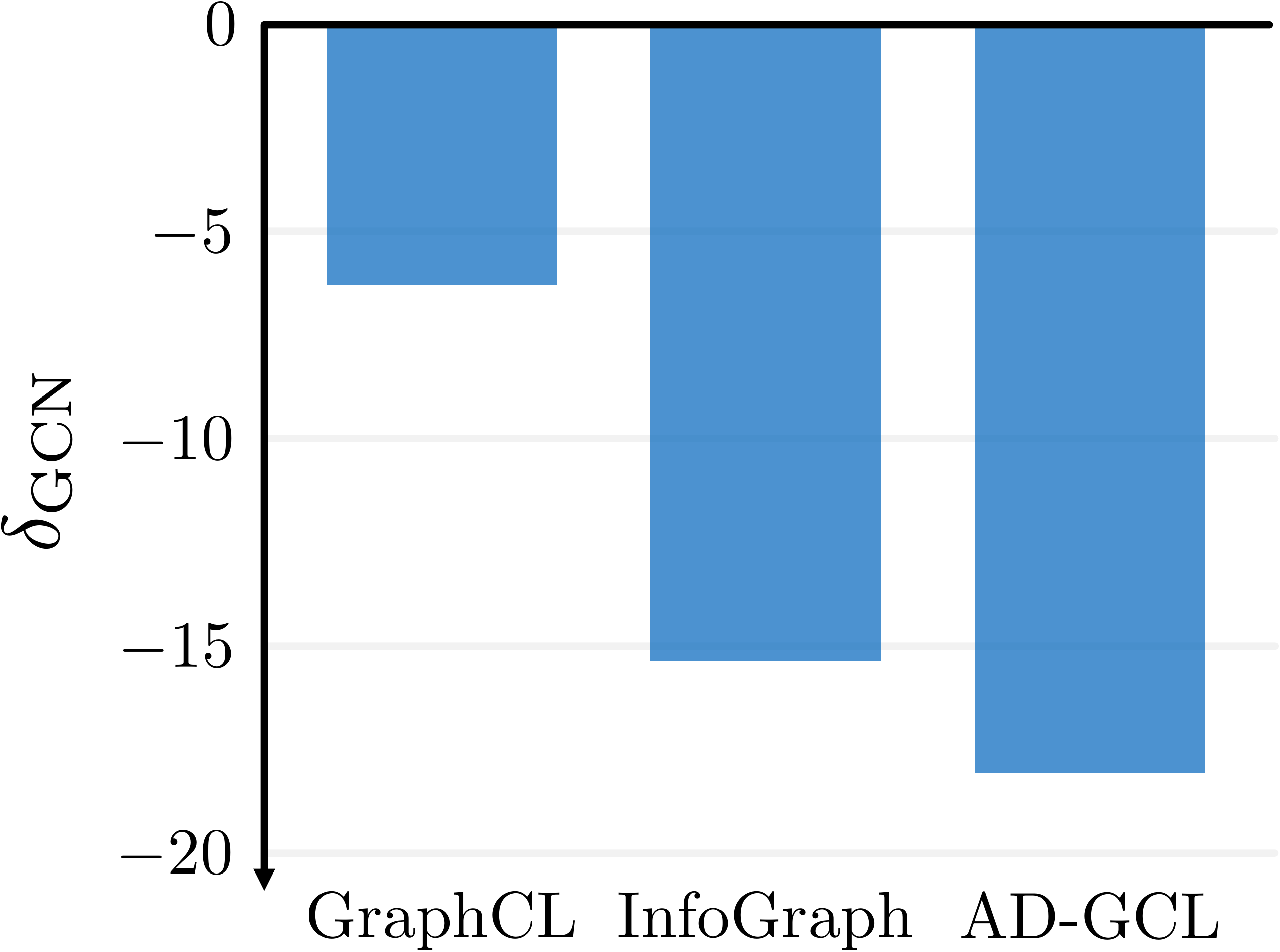}
    \caption{Average robustness improvement of graph contrastive learning methods w.r.t.\ the GCN baseline, across graph classification datasets. All assessed contrastive methods seem to \textit{lower the robustness} in \textit{evasion} scenarios, hence making the models more susceptible to adversarial attacks (see \cref{tab:res_gc}, and \cref{eq:model_delta}).} \label{fig:first_page}
    \vspace{-1.2em}
\end{wrapfigure}

Contrastive learning (CL) \citep{hadsell2006dimensionality, bachman2019learning, chen2020simple, jaiswal2021survey, liu2021selfsupervised, khan2022contrastive} has emerged as a powerful framework within the field of self-supervised representation learning. Contrastive learning is primarily applied to text and images, and aims at learning latent representations by leveraging information inherent to the data. In contrastive learning, a common strategy involves generating multiple \textit{augmentations} or views of an input entity and training the network to maximize the mutual information across these views \citep{bachman2019learning}. This approach pushes the network to learn latent representations that are invariant to \textit{data perturbations} similar to the augmentations used during training \citep{bachman2019learning}. Notably, the choice of augmentations plays a pivotal role in influencing the performance of the model on downstream predictive tasks \citep{chen2020simple}.

Despite not consistently matching the performance of fully supervised learning (SL) methods, CL exhibits remarkable potential in improving model robustness, as observed in previous works \citep{hendrycks2019usinga, shi2022how}. This advantage is especially pronounced when compared to the \textit{target-driven} objective of SL methods. The benefits introduced by CL extend to various aspects of robustness, including robustness to adversarial examples \citep{kim2020adversarial}, label corruptions \citep{xue2022investigating}, and distribution shifts \citep{shi2022how}. 

While extensive research has investigated and demonstrated the advantages of CL in computer vision and natural language processing domains, its effectiveness in the realm of graph-structured data remain uncertain. Only recently, graph contrastive learning (GCL) methods have emerged to address this challenge \citep{velickovic2018graph, sun2019infograph, you2020graph, zhu2020deep, zhu2021graph, xie2022selfsupervised}. Analogous to CL, the choice of augmentations in GCL profoundly influences model accuracy, and researchers have introduced novel augmentation strategies to enhance the performance of GCL models on downstream tasks \citep{li2022disentangled, suresh2021adversarial, wang2022augmentationfree}.

However, despite similarities with CL and the apparent impact of augmentations, the extent to which augmentations contribute to enhancing the robustness of learned representations in GCL remains an open question. Furthermore, works aiming to improve robustness by employing adversarially crafted augmentations often lack robustness evaluations \citep{suresh2021adversarial}. Those that do assess robustness \citep{jovanovic2021robust, li2023effective} typically rely on simplistic proxies, such as random noise or transfer attacks \citep{zugner2018adversarial, zugner2019adversarial}, which can potentially overestimate actual robustness \citep{mujkanovic2022are}. Therefore, there exists a research gap necessitating a consistent evaluation of the robustness of GCL methods to adversarial attacks.

In this paper, we introduce the \textit{robustness evaluation protocol} \pcref{fig:pipeline} designed to empirically assess the robustness of various GCL methods. These methods are evaluated against \textit{adaptive} adversarial attacks that target the structure of graphs. Our focus include both node and graph classification tasks, particularly in the \textit{evasion} (test time) scenario. We present our findings across multiple real-world datasets and diverse attack strategies. It is important to note that comparing the robustness of two models using an attack is a challenging task, as it essentially provides upper bounds on the worst-case perturbed robustness. However, we posit that these upper bounds provide valuable insights into the relative robustness of GCL models, even if they do not capture the entire landscape of robustness scenarios.
Through extensive empirical analysis, our goal is to asses the efficacy of GCL methods in adversarial scenarios, thereby contributing to a deeper understanding of their practical utility and limitations in real-world applications. Our investigations shows the inadequacy of naive proxies often used to assess the robustness of GCL methods, highlighting how they tend to overestimate actual robustness under more realistic and adaptive attack conditions.
    \section{Background}

\paragraph{Notation.}
We define \smash{$G = (V, E)$} as an undirected graph, where \smash{$V$} represents the set of nodes with \smash{$n = |V|$} nodes, and \smash{$E$} represents the set of edges with \smash{$m = |E|$} edges. In this representation, we allow for a node features matrix \smash{$\mX \in \R^{n \times f}$} and denote the graph structure through a symmetric adjacency matrix \smash{$\mA \in \{0,1\}^{n \times n}$}. Here, \smash{$\mA_{ij} = 1$} if there exists an edge between nodes \smash{$i$} and \smash{$j$}, and \smash{$\mA_{ij} = 0$} otherwise. For node-level tasks within a given graph \smash{$G$}, our objective is to learn the latent representation \smash{$\vh_v$} for each node \smash{$v \in V$}. Instead, for graph-level tasks involving a collection of graphs \smash{$\gG = \{G_1, G_2, \dots\}$}, our goal is to learn the latent representation \smash{$\vh_G$} for each graph \smash{$G \in \gG$}. We let $\gP$ denote the distribution of unlabeled graphs over the input space $\gG$. To maintain simplicity and clarity throughout this paper, we consistently refer to both node and graph representations as \smash{$\vh$}.

\paragraph{Graph contrastive learning.}
Graph contrastive learning (GCL) has emerged as a framework specifically designed to train graph neural networks (GNNs) \citep{scarselli2009graph, kipf2016semisupervised, gilmer2017neural, bronstein2017geometric} in a self-supervised manner. The fundamental concept behind GCL involves generating augmented views of an input graph and maximizing the \textit{agreement} between these views pertaining to the same node or graph. This agreement is commonly quantified by the mutual information \smash{$\gI(\vh_i, \vh_j) = D_{KL}(p(\vh_i, \vh_j) \| p(\vh_i)p(\vh_j))$} between a pair of representations $\vh_i$ and $\vh_j$, where \smash{$D_{KL}$} represents the Kullback-Leibler divergence \citep{kullback1951on}. Contrastive learning aims to maximize the mutual information between two views treated as random variables. Specifically, it trains the encoders to be pull together representations of positive pairs drawn from the joint distribution \smash{$p(\vh_i, \vh_j)$} and push apart representations of negative pairs derived from the product of marginals \smash{$p(\vh_i)p(\vh_j)$}. To computationally estimate and maximize mutual information in contrastive learning, three lower bounds are commonly used \citep{hjelm2018learning}: Donsker-Varadhan (DV) \citep{donsker1975asymptotic}, Jensen-Shannon (JS) \citep{nowozin2016fgan}, and noise-contrastive estimation (InfoNCE) \citep{gutmann2010noise}. A mutual information estimation is usually computed based on a discriminator $\gD : \R^q \times \R^q \to \R$ that maps the representations of two views to an agreement score between them. In particular, given two representations $\vh_i$ and $\vh_j$ computed from the graph $G = (\mA, \mX)$ and a discriminator $\gD$, the Jensen-Shannon estimator is defined as follows:
\begin{equation*}
    \hat{\gI}^{(\text{JS})}(\vh_i, \vh_j)= \E_{G \sim \gP} [\log(\gD(\vh_i, \vh_j))]+ \E_{[G, G'] \sim \gP \times \gP} [\log(1 - \gD(\vh_i, \vh_j'))],
\end{equation*}
while the InfoNCE estimator can be written as:
\begin{equation*}
    \hat{\gI}^{(\text{NCE})}(\vh_i, \vh_j)=\E_{[G, \mK] \sim \gP \times \gP^N} \left[\log\frac{e^{\gD(\vh_i, \vh_j)}}{\sum_{G' \in \mK}e^{\gD(\vh_i, \vh_j')}}\right] + \log N,
\end{equation*}
where $N$ is the number of negative samples, and $\mK$ consists of $N$ i.i.d. graphs sampled from $\gP$. In practice we compute the InfoNCE over mini-batches of size $N + 1$ and minimize the following loss:
\begin{equation}
    \label{eq:L_cntr}
    \begin{split}
        \gL_\text{InfoNCE} = -\frac{1}{N+1} \sum_{G \in \gB}\left[\log\frac{e^{\gD(\vh_i, \vh_j)}}{\sum_{G' \in \gB / \{G\}}e^{\gD(\vh_i, \vh_j')}}\right],
    \end{split}
\end{equation}
where \smash{$G \in \gB$} is a graph in the mini-batch \smash{$\gB$}, and \smash{$\vh_i, \vh_j$} are representations computed from the graph. Intuitively, the optimization of \cref{eq:L_cntr} aims at making the agreement between the representations $\vh_i$ and $\vh_j$ of views from the same graph $G$ higher, while decreasing the agreement between the representation $\vh_i$ of a view of graph $G$ and the representation $\vh_j'$ of a view from the $N$ negative samples \smash{$\gB / \{G\}$}. A common contrastive loss that follows this principle is the NT-Xent loss \citep{sohn2016improved, wu2018unsupervised, oord2018representation}, which uses \smash{$\gD = \langle g(\vh_i), g(\vh_j) \rangle /\tau$}, where \smash{$g: \R^q \to \R^s$} projects the representations to a lower dimensional space, $\tau$ is a temperature parameter, and \smash{$\langle \cdot, \cdot \rangle$} represents the dot product.

For a description of the models considered in our evaluation, please refer to \cref{app:models}. We also refer the reader to \citet[and references therein]{xie2022selfsupervised} for a comprehensive introduction to GCL.

\paragraph{Adversarial robustness.} 
In the field of machine learning (ML), adversarial examples have become a significant concern \citep{goodfellow2014explaining}. Adversarial examples are perturbed inputs that, albeit being indistinguishable from the original input, lead to a change in the prediction of the model. Adversarial robustness \citep{carlini2017evaluating, madry2018deep}, therefore, refers to the capability of an ML model to withstand such attacks and maintain accuracy, especially in important safety-critical applications. Various techniques have been developed to enhance adversarial robustness, including adversarial training, feature engineering, gradient masking, and ensemble methods \citep{chakraborty2018adversarial}. Notably, contrastive learning has emerged as a promising approach to improve model robustness \citep{kim2020adversarial}.

While much of the research has been focused on traditional data types like images and text, there is a growing interest in studying and enhancing the robustness of ML models applied to graph data \citep{dai2018adversarial, gunnemann2022graph}. Graphs are used to represent complex relationships, and ensuring their robustness is critical in domains such as social network analysis, recommendation systems, and fraud detection. Addressing adversarial attacks in graph-based ML models is an evolving area, with methods tailored to graph structures, including node and edge perturbations \citep{zugner2018adversarial, bojchevski2019adversarial}, being actively explored to secure these systems.

Refer to \cref{app:attacks} for a comprehensive description of the attacks considered in our evaluation. We also refer the reader to \citet{gunnemann2022graph} for an introduction to adversarial robustness on graphs.
    \section{Evaluating adversarial robustness of graph contrastive learning methods}

\begin{figure}[t]
    \centering
    \includegraphics[width=0.8\linewidth]{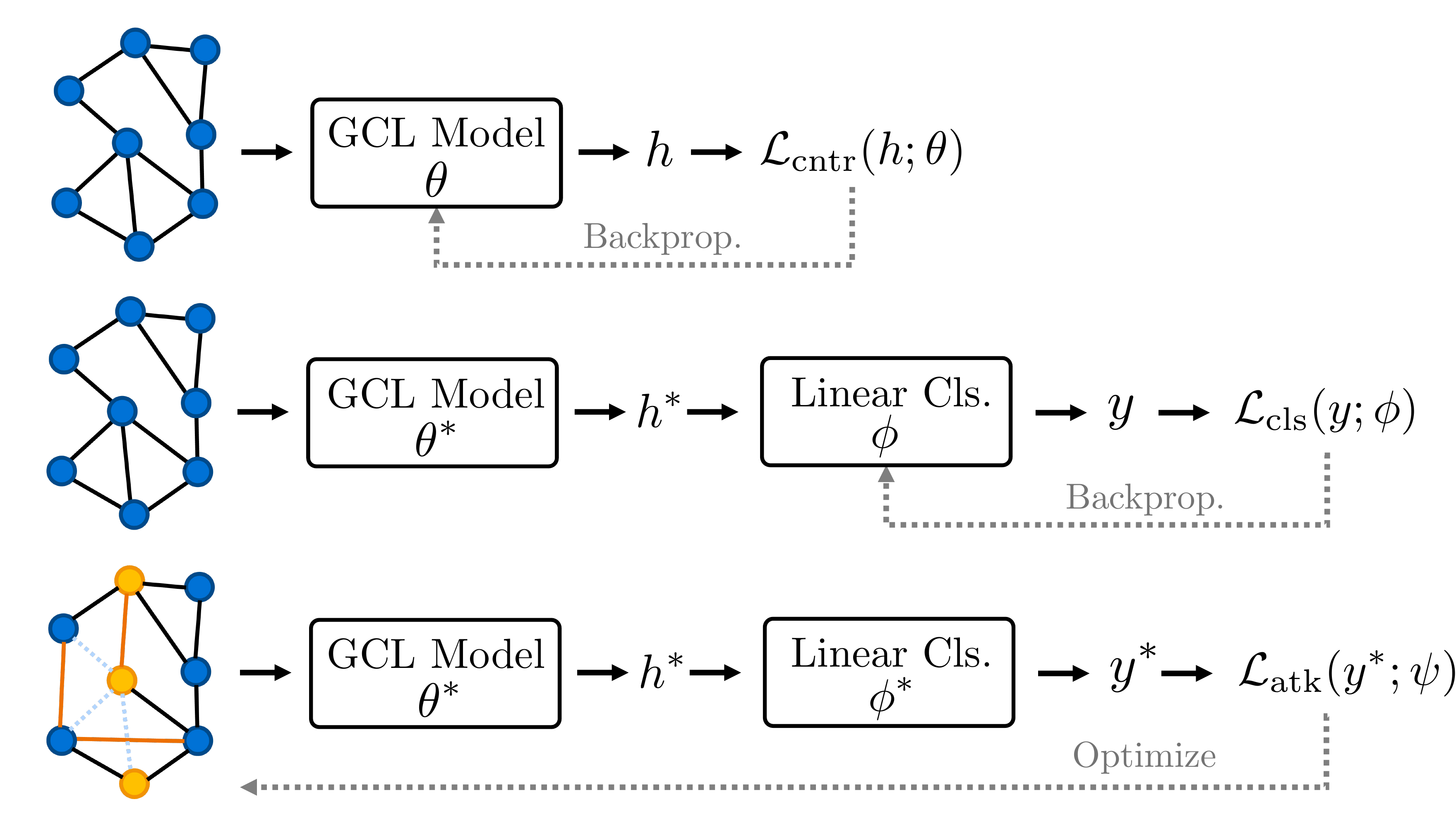}
    \caption{The three steps of our proposed \textit{robustness evaluation protocol} for graph contrastive learning methods: (1) encoder training, (2) linear classifier training, (3) evasion attack.}
    \label{fig:pipeline}
\end{figure}

\paragraph{Problem setup.}
Our focus is on evaluating the adversarial robustness w.r.t.\ both structural and feature-based attacks on graph neural networks (GNNs) trained using graph contrastive learning methods for node and graph classification tasks. Specifically, we assess adversarial robustness during test time, i.e., \textit{evasion} attacks. 

\subsection{Robustness evaluation protocol}
\label{sec:rob_eval_protocol}
Inspired by the linear evaluation protocol proposed by \citet{velickovic2019deep}, we introduce our \textit{robustness evaluation protocol} as illustrated in \cref{fig:pipeline}. This protocol establishes a general yet consistent framework for assessing the robustness of self-supervised (or unsupervised) models against adversarial attacks across various attack schemes. The proposed robustness evaluation protocol can be applied to both node and graph classification tasks and is adaptable to different loss functions employed in the self-supervised encoder, linear classifier, and adversarial attack, respectively.

This evaluation protocol serves as a foundational structure that enables standardized and comprehensive evaluations of model robustness. By encompassing various loss functions and attack types, it ensures that the assessment is applicable to a wide range of scenarios. The evaluation consists of three steps, as described in the following paragraphs.

\paragraph{(Step 1) Encoder training.}
The initial step of the evaluation protocol is the training of the GCL encoder. The encoder, \smash{$f_\theta: \R^{n \times f} \times \{0,1\}^{n \times n} \to \gH$}, takes an input graph characterized by its node attributes and adjacency matrix, and outputs latent representations \smash{$\vh \in \gH$} for either the entire graph or its individual nodes, depending on the specific task it is trained on. Training the encoder is a critical phase where the model learns to capture meaningful graph structure and representation. This phase is guided by minimizing a contrastive loss function $\gL_\text{cntr}(\vh; \theta)$, as the one defined in \cref{eq:L_cntr}. The objective is to ensure that similar instances in the graph result in representations that are close in the latent space, while dissimilar instances are pushed apart. 

\paragraph{(Step 2) Linear classifier training.}
In this phase, we train a linear classifier, $g_\phi: \gH \to \gY$, that takes as input the latent representations $\vh^*$ generated by the encoder $f_{\theta^*}$ with learned \textit{fixed} parameters $\theta^*$, and outputs the downstream predictions $\vy \in \gY$. In the classification setting, the classifier is trained by minimizing the cross-entropy loss:
\begin{equation}    
\gL_\text{cls} = -\sum_{i=1}^{n} \sum_{j=1}^{|\gY|} \bar{y}_{ij} \log(y_{ij}), 
\end{equation}
where $\bar{y}_{ij}$ represents the ground truth label for node $i$ with respect to class $j$, and $y_{ij}$ is the predicted probability assigned by the classifier to node $i$ belonging to class $j$. It is important to highlight that the parameters $\theta^*$ of the encoder are kept fixed during this phase, and only the parameters of the linear classifier $\phi$ are updated, hence preserving the self-supervised nature of the encoder.

\paragraph{(Step 3) Adversarial attack.}
The last step of the evaluation protocol consists of attacking the encoder and the linear classifier. We perturb the input graph $G = (\mX, \mA)$ by approximating:
\begin{equation}\label{eq:adv_atk}
    \tilde{G} = (\tilde{\mX}, \tilde{\mA}) = \arg\min\nolimits_{\tilde{G} \in \varphi(G)} \gL_\text{atk}(g_{\phi^*}(f_{\theta^*}(\tilde{\mX}, \tilde{\mA}))),
\end{equation}
where $\varphi(G)$ defines the admissible perturbations that the attack optimizes over. Following the established literature on adversarial robustness of GNNs, we focus on perturbations related to the structure of the graphs: \smash{$\varphi(G) = \{\tilde{G} \,|\, \|\tilde{\mA} - \mA\|_0 \le 2 \Delta \,\land\, \tilde{\mA}^\top = \tilde{\mA}\}$}. This choice of \smash{$\varphi(G)$} permits up to \smash{$\Delta$} edge additions/removals while ensuring symmetry in \smash{$\tilde{\mA}$}, hence, that the perturbed graph is still undirected. In our study, we keep the parameters $\theta^*$ of the $f_{\theta^*}$ and the parameters $\phi^*$ of the $g_{\phi^*}$ \textit{fixed} (evasion scenario). However, our framework naturally extends to training-time attacks (poisoning).  
We employ various attack methods, including random edge flips as a baseline, Projected Gradient Descent (PGD) \citep{xu2019topology}, and two variants of Randomized Block Coordinate Descent (R-BCD) \citep{geisler2021robustness}. Furthermore, we use these attacks as \textit{global} attacks, i.e., they attack the graph-level prediction or target the prediction of all test nodes jointly. This is in contrast to Nettack \citep{zugner2018adversarial}, which is a \textit{local} attack targeting individual nodes. 

An important distinction in our evaluation, as opposed to prior GCL work, is that the PGD and R-BCD attacks are \emph{adaptive} \citep{mujkanovic2022are}. That is, the attacks are fully white-box and capable of adapting to the different representations learned by GCL models, as opposed to supervised graph learning. As we will see in our empirical evaluation, employing adaptive and global attacks for assessment raises questions about the robustness benefits claimed by many GCL methods.

\subsection{Measuring Adversarial Robustness}
\label{sec:rel_drop}
In machine learning, particularly in the context of evaluating adversarial robustness across diverse datasets and models, it is crucial to establish standardized metrics for comparative analysis. 

\paragraph{Relative Adversarial Accuracy Drop.} We introduce a new metric that provides a standardized approach for assessing adversarial robustness, enabling meaningful comparisons across different datasets and machine learning models, the \textit{Relative Adversarial Accuracy Drop}, defined as follows:
\begin{equation}
    \label{eq:acc_drop}
    \relDrop = \gR(\text{acc}_\text{clean}, \text{acc}_\text{adv}) = \frac{\text{acc}_\text{clean} - \text{acc}_\text{adv}}{\text{acc}_\text{clean}}.
\end{equation}
$\relDrop$ quantifies the relative reduction in accuracy between clean ($\text{acc}_\text{clean}$) and adversarial ($\text{acc}_\text{adv}$) predictions. The normalization by $\text{acc}_\text{clean}$ confines $\gR$ within the interval $[0, 1]$, ensuring fair comparisons, even when models exhibit varying levels of performance on clean data. Our primary objective is to evaluate the robustness of the models rather than their absolute performance on a specific dataset. A lower value of $\gR$ indicates higher robustness against adversarial attacks, signifying a smaller decrease in accuracy when exposed to adversarial inputs compared to clean data.

\paragraph{Model comparison.} When comparing a model $m$ to a \textit{reference} model $r$ across various datasets $\gD = \{D_i\,|\,i=1, \dots\}$\footnote{For instance, $\gD = \{\textsc{PROTEINS}, \textsc{NCI1}, \textsc{DD}\}$.}, we employ the following metric:
\begin{equation}
\label{eq:model_delta}
\delta^m_r = \frac{1}{|\gD|} \sum_{D \in \gD}(\gR^D_m - \gR^D_r).
\end{equation}
Here, $\gR^D_m$ and $\gR^D_r$ represent the $\relDrop$ of model $m$ and the reference model $r$ on dataset $D$, respectively. A positive $\delta^m_r$ indicates that model $m$ consistently outperforms the reference model across the datasets, while a negative $\delta^m_r$ suggests the opposite. Therefore, the $\delta^m_r$ serves as a comprehensive measure of the overall comparative performance of the model of interest with respect to the reference model across diverse datasets.

    \section{Empirical Evaluation}

Our empirical evaluation encompasses a comprehensive analysis involving multiple node and graph classification datasets, as well as a variety of graph contrastive learning (GCL) models. We measure both clean accuracy and perturbed accuracy under various \textit{static} and \textit{adaptive} attack scenarios and report the relative drop in accuracy compared to the clean accuracy.

\subsection{Setup}
In our evaluation, we use the model architectures as reported in the reference implementations. This includes the type and sequence of layers, choice of activation functions, and the application of dropout. Given that our primary objective is to assess the extent to which GCL methods are affected by adversarial attacks, we do \textit{not} perform hyperparameter tuning. Instead, we employ the best hyperparameters for each model as documented in their original implementations. We utilize the Adam optimizer \citep{kingma2014adam} across all models, and set the learning rate as defined in the reference papers (see \cref{app:hparams}). Our results are averaged over 15 different initializations of the GCL encoder and linear classifier. 
We follow the robustness evaluation protocol, introduced in \cref{sec:rob_eval_protocol}, for all contrastive-based models. For non-contrastive models, such as GCN \citep{kipf2016semisupervised} and GIN \citep{xu2018how}, we train the network in a supervised manner, and then apply Step 3 of the protocol. Additionally, we compute the \textit{Relative Adversarial Accuracy Drop} \smash{$\relDrop$} \pcref{sec:rel_drop} and identify the minimum value across different attack schemes.

\paragraph{Graph classification.} 
In the context of graph classification, we are given a dataset \smash{$\gD = \{(G_i, y_i)\,|\,i = 1, \dots, g\}$} that comprises a collection of $g$ graphs $G_i$ along with their respective labels $y_i$. The central objective of this task is to assign \textit{entire} graphs to specific classes. To evaluate the performance of GCL models in this task, we employ three benchmark datasets: PROTEINS, NCI1, and DD \citep{morris2020tudataset}. \cref{tab:datasets_gc} in \cref{app:results} reports detailed description of the dataset statistics. To ensure a consistent evaluation process across all datasets, we employ a random data split strategy, allocating 80\% of the nodes for training and reserving the remaining 20\% for testing. We perform mini-batch training with batch size of 64.
\begin{table}[t]
\centering
\caption{\textbf{Graph classification} performance: clean and perturbed accuracy mean $\pm$ std., and $\relDrop$ (percentage), averaged over 15 runs with different weight initializations. The results include the accuracy for different contrastive and non-contrastive models, and different datasets under the attack scheme that most influences accuracy. The attacks are allowed (\(\Delta\)) to change 5\% of edges. Models achieving the three highest relative drop are highlighted as \first{first}, \second{second}, and \third{third}. The higher the worse.}
\label{tab:res_gc}
\resizebox{0.82\linewidth}{!}{%
\begin{tabular}{@{}llcccccc@{}}
\toprule
\textbf{Model} &  & \multicolumn{2}{c}{\textsc{PROTEINS}} & \multicolumn{2}{c}{\textsc{NCI1}} & \multicolumn{2}{c}{\textsc{DD}} \\ \midrule
\multirow{2}{*}{GCN}
& \textsc{Clean} & 
\scriptsize{$73.98 \pm 1.00$} & \multirow{2}{*}{$\downarrow ~~8.04$} &
\scriptsize{$74.67 \pm 0.80$} & \multirow{2}{*}{$\downarrow 54.59$}  & 
\scriptsize{$70.02 \pm 0.54$} & \multirow{2}{*}{$\downarrow \first{87.57}$} \\
& \textsc{Attack} & 
\scriptsize{$68.04 \pm 0.90$} & & 
\scriptsize{$33.91 \pm 1.36$} & &  
\scriptsize{$~~8.58 \pm 7.35$} & \\
\midrule
\multirow{2}{*}{GIN}
& \textsc{Clean} & 
\scriptsize{$66.02 \pm 4.60$} & \multirow{2}{*}{$\downarrow 30.24$} & 
\scriptsize{$76.04 \pm 0.91$} & \multirow{2}{*}{$\downarrow 49.17$} &
\scriptsize{$63.44 \pm 3.19$} & \multirow{2}{*}{$\downarrow \third{74.51}$} \\
& \textsc{Attack} & 
\scriptsize{$46.05 \pm 6.86$} & & 
\scriptsize{$38.65 \pm 2.48$} & &   
\scriptsize{$16.17 \pm 6.16$} & \\
\midrule
\multirow{2}{*}{InfoGraph}
& \textsc{Clean} & 
\scriptsize{$63.93 \pm 6.43$} & \multirow{2}{*}{$\downarrow \second{50.83}$} & 
\scriptsize{$66.87 \pm 3.72$} & \multirow{2}{*}{$\downarrow \second{78.10}$} & 
\scriptsize{$64.77 \pm 2.41$} & \multirow{2}{*}{$\downarrow 68.92$}\\
& \textsc{Attack} & 
\scriptsize{$31.53 \pm 7.67$} & &  
\scriptsize{$14.71 \pm 4.13$} & &  
\scriptsize{$20.11 \pm 4.24$} & \\
\midrule
\multirow{2}{*}{GraphCL}
& \textsc{Clean} & 
\scriptsize{$65.93 \pm 4.02$} & \multirow{2}{*}{$\downarrow \third{38.89}$} & 
\scriptsize{$73.09 \pm 3.50$} & \multirow{2}{*}{$\downarrow \third{55.94}$} & 
\scriptsize{$68.85 \pm 4.83$} & \multirow{2}{*}{$\downarrow \second{75.34}$}\\
& \textsc{Attack} & 
\scriptsize{$40.29 \pm 9.61$} & & 
\scriptsize{$32.20 \pm 3.48$} & &  
\scriptsize{$15.74 \pm 7.66$} & \\
\midrule
\multirow{2}{*}{AD-GCL}
& \textsc{Clean} & 
\scriptsize{$73.66 \pm 1.36$} & \multirow{2}{*}{$\downarrow \first{64.28}$} & 
\scriptsize{$71.79 \pm 0.94$} & \multirow{2}{*}{$\downarrow \first{80.01}$} & 
\scriptsize{$76.71 \pm 2.07$} & \multirow{2}{*}{$\downarrow 61.71$} \\
& \textsc{Attack} & 
\scriptsize{$26.31 \pm 4.33$} & & 
\scriptsize{$14.35 \pm 1.34$} & &  
\scriptsize{$29.17 \pm 5.27$} & \\
\bottomrule
\end{tabular}
}
\end{table}

\paragraph{Node classification.}
In semi-supervised node classification, we are given a graph \smash{$G = (V, E)$}, with \smash{$V=\{v_i\,|\,i=1,\dots, n\}$} and we assume only a subset of nodes \smash{$\Tilde{V} \subset V$} are labeled, with node labels $y_j$ for \smash{$j \in \Tilde{V}$}. The aim is to assign unlabeled nodes \textit{within} a graph to specific classes. We conduct experiments on well-established node classification datasets, namely Cora, Citeseer \cite{sen2008collective}, Pubmed \citep{namata2012querydriven}. We also include in our analysis the \textit{large-scale} graph OGB-arXiv \citep{hu2020open} (see \cref{tab:datasets_nc} for a description of the dataset statistics).
We adhere to the data splits proposed in \citep{yang2016revisiting} for Cora, Citeseer, and Pubmed, while opting for a random split in the case of OGB-arXiv. The random split allocates 80\% of nodes for training and reserves the remaining 20\% for testing. We perform full-batch training, using all nodes in the training set at every epoch.

\subsection{Discussion}
The results for graph and node classification tasks are presented in \cref{tab:res_gc,tab:res_nc}. These tables report the model accuracies under two conditions: (\textit{i}) clean, unperturbed data and (\textit{ii}) data subjected to various adversarial attack schemes. Specifically, we report the minimum accuracy achieved across different attack schemes for each dataset and model. A more detailed report of the results across various attack schemes can be found in \cref{tab:res_gc_complete,tab:res_nc_complete}. We emphasize the models that exhibit the worst three $\relDrop$ values.

In our analysis of graph classification, we observe that the GCL methods used in our evaluation (InfoGraph \citep{sun2019infograph}, GraphCL \citep{you2020graph}, AD-GCL \citep{suresh2021adversarial}) do not succeed in enhancing adversarial robustness. Notably, on PROTEINS and NCI1 datasets, these three GCL models consistently achieve the first, second, and third worst results, indicating a significant drop in accuracy under adversarially perturbed data compared to non-contrastive models like GCN and GIN. The DD dataset, however, shows slightly different results, with GCN being the weakest model in terms of robustness, while GraphCL, a GCL model, takes the second position.

Moving to node classification experiments, the results are mixed. GCL models (DGI \citep{velickovic2018graph}, GraphCL \citep{you2020graph}, GCA \citep{zhu2021graph}) generally perform as the worst or second-worst models across Cora, Citeseer, and Pubmed datasets. However, they show improved robustness when compared to GCN on the large-scale OGB-arXiv dataset. An interesting observation is that GCA goes out of memory (OOM) on the OGB-arXiv dataset due to the high complexity of loss computation. It is worth noting that DGI consistently performs the best across all datasets. This behavior might be attributed to DGI's particular training strategy, which involves creating corruptions of the original graph and forcing the network to recognize them as different, as opposed to many contrastive methods that are based on identifying different views as similar. A more in-depth analysis of this behavior remains a direction for future research.

When analyzing the impact of \textit{adaptive} versus \textit{static} attacks, and \textit{global} versus \textit{local} attacks, it becomes evident from both \cref{tab:res_gc_complete} and \cref{tab:res_nc_complete} that adaptive attacks generate perturbed graph structures that are more detrimental to every model. This reinforces our argument that static attacks, such as random flipping, or local attacks like Nettack \citep{zugner2018adversarial}, are insufficient for evaluating the robustness of a method.

In summary, our findings indicate that in both graph and node classification tasks, GCL methods do not show a clear advantage in terms of improving the adversarial robustness of graph neural networks. In some instances, contrastive training can even lead to a further deterioration in performance.

\begin{table}[t]
\centering
\caption{\textbf{Node classification} performance: clean and perturbed accuracy mean $\pm$ std., and $\relDrop$ (percentage), averaged over 15 runs with different weight initializations. The results include the accuracy for different contrastive and non-contrastive models, and different datasets under the attack scheme that most influences accuracy. The attacks are allowed (\(\Delta\)) to change 5\% of edges. Models achieving the three highest relative drop are highlighted as \first{first}, \second{second}, and \third{third}. The higher the worse.}
\label{tab:res_nc}
\resizebox{\linewidth}{!}{%
\begin{tabular}{@{}llcccccccc@{}}
\toprule
\textbf{Model} & & \multicolumn{2}{c}{\textsc{Cora}} & \multicolumn{2}{c}{\textsc{Citeseer}} & \multicolumn{2}{c}{\textsc{Pubmed}} & \multicolumn{2}{c}{\textsc{OGB-arXiv}} \\ \midrule
\multirow{2}{*}{GCN}
& \textsc{Clean} & 
\scriptsize{$77.57 \pm 1.11$} & \multirow{2}{*}{$\downarrow \second{23.50}$} &
\scriptsize{$63.99 \pm 1.27$} & \multirow{2}{*}{$\downarrow \second{25.46}$} &
\scriptsize{$75.31 \pm 0.80$} & \multirow{2}{*}{$\downarrow \third{24.31}$}  &
\scriptsize{$52.39 \pm 0.79$} & \multirow{2}{*}{$\downarrow \first{82.67}$}  \\
& \textsc{Attack} & 
\scriptsize{$59.35 \pm 1.70$} & & 
\scriptsize{$47.69 \pm 2.06$} & & 
\scriptsize{$57.01 \pm 1.28$} & &
\scriptsize{$~~9.08 \pm 2.10$} & \\
\midrule
\multirow{2}{*}{DGI}
& \textsc{Clean} & 
\scriptsize{$83.24 \pm 1.37$} & \multirow{2}{*}{$\downarrow ~~9.08$} &
\scriptsize{$72.91 \pm 1.99$} & \multirow{2}{*}{$\downarrow ~~7.47$} &
\scriptsize{$81.46 \pm 0.79$} & \multirow{2}{*}{$\downarrow ~~5.23$} &
\scriptsize{$60.18 \pm 0.72$} & \multirow{2}{*}{$\downarrow \third{11.45}$} \\
& \textsc{Attack} & 
\scriptsize{$75.69 \pm 1.86$} & & 
\scriptsize{$67.46 \pm 2.13$} & & 
\scriptsize{$77.19 \pm 0.92$} & &
\scriptsize{$53.28 \pm 0.69$} & \\
\midrule
\multirow{2}{*}{GraphCL}
& \textsc{Clean} & 
\scriptsize{$71.99 \pm 1.35$} & \multirow{2}{*}{$\downarrow \first{24.00}$} &
\scriptsize{$59.57 \pm 1.45$} & \multirow{2}{*}{$\downarrow \third{21.89}$} &
\scriptsize{$74.29 \pm 1.75$} & \multirow{2}{*}{$\downarrow \second{27.33}$} &
\scriptsize{$52.32 \pm 0.22$} & \multirow{2}{*}{$\downarrow \second{72.65}$} \\
& \textsc{Attack} & 
\scriptsize{$54.71 \pm 1.46$} & & 
\scriptsize{$46.53 \pm 1.87$} & & 
\scriptsize{$53.99 \pm 1.77$} & &
\scriptsize{$14.31 \pm 0.19$}\\
\midrule
\multirow{2}{*}{GCA}
& \textsc{Clean} & 
\scriptsize{$79.07 \pm 1.36$} & \multirow{2}{*}{$\downarrow \third{22.08}$} &
\scriptsize{$60.14 \pm 2.24$} & \multirow{2}{*}{$\downarrow \first{27.51}$} &
\scriptsize{$78.62 \pm 0.98$} & \multirow{2}{*}{$\downarrow \first{27.80}$} &
\multicolumn{2}{c}{\multirow{2}{*}{OOM}} \\
& \textsc{Attack} & 
\scriptsize{$61.63 \pm 3.18$} & & 
\scriptsize{$43.61 \pm 2.13$} & & 
\scriptsize{$56.77 \pm 1.69$} & & 
\multicolumn{2}{c}{} \\
\bottomrule
\end{tabular}
}
\end{table}
    \section{Conclusion}

We thoroughly evaluated the robustness of graph contrastive learning (GCL) methods to adaptive adversarial attacks on graph structures. Our investigation included node and graph classification tasks across multiple real-world datasets and various attack strategies.

Our results show that GCL methods do \textit{not} consistently exhibit improved adversarial robustness compared to non-contrastive methods. In specific datasets and attack scenarios, GCL models perform even worse in robustness. This finding challenges the common belief that CL methods, successful in other domains, automatically translate to enhanced robustness in graph-structured data. 

One notable discovery is the varying performance of GCL models, with DGI consistently outperforming others in node classification tasks. This behavior may be attributed to DGI's unique training strategy, differentiating between corruptions and the original graph instead of maximizing mutual information between augmentations. This aspect deserves further investigation to comprehend its implications for adversarial robustness.

In conclusion, our findings suggest that while GCL methods hold promise for representation learning on graph-structured data, they do not inherently guarantee improved adversarial robustness. The effectiveness of GCL methods in adversarial scenarios is nuanced and context-dependent, necessitating further research to uncover the precise conditions under which they excel or falter. Exploring additional models, datasets, and a broader range of adversarial attacks should be considered in future research to comprehensively understand GCL robustness in practice.

    \section*{Acknowledgments}
This work was supported by the German Federal Ministry of Education and Research (BMBF) (HOPARL, 031L0289C). The authors of this work take full responsibility for its content.
    \bibliography{references}
    \newpage
    \section*{Appendix}
    \appendix
    \section{Graph contrastive learning models}
\label{app:models}

\textbf{Deep graph infomax (DGI)} \citep{velickovic2018graph} is a self-supervised approach for learning node representations in graph-structured data. It aims to maximize mutual information between patch representations and high-level graph summaries. DGI does not rely on random walk objectives, making it suitable for both transductive and inductive learning.

In the single-graph setup, the authors propose to corrupt the input graph $(\mX, \mA)$ with a \textit{corruption function} $\gC$, and obtain a negative sample \smash{$(\tilde{\mX},\tilde{\mA}) \sim \gC(\mX,\mA)$}. They then use a trainable encoder $\gE$ for both clean and corrupted input to obtain patch representations \smash{$\vh_i$} and \smash{$\tilde{\vh}_i$} for each node $i$ in the clean and corrupted graph, respectively: \smash{$\mH = \gE(\mX, \mA) = \{\vh_1, \dots, \vh_N\}$}, and \smash{$\tilde{\mH} = \gE(\mX, \mA) = \{\tilde{\vh}_1, \dots, \tilde{\vh}_M\}$}. The encoder is a one-layer graph convolutional network (GCN) \citep{kipf2016semisupervised}, but the framework is general enough to use different models.The patch representations of the input graph are then used to create a summary of the graph itself \smash{$\vs = \gR(\mH) = \sigma((\sum_{i=0}^{N}\vh_i)/N)$}, where $\gR$ is a non-linear readout function. Patch representations coming from the input graph and those coming from the corrupted graph are then passed through a discriminator function $\gD$ that scores summary-patch representation pairs by applying a simple bilinear scoring function. Scores are then converted to probabilities of $(\vh_i, \vs)$ being a positive example, and \smash{$(\tilde{\vh}_i, \vs)$} being a negative example. The whole network is trained end-to-end via cross-entropy, with label $+1$ assigned to positive examples, and $-1$ assigned to negative ones.

\textbf{InfoGraph} \citep{sun2019infograph} learns graph-level representations by maximizing the mutual information between the representation of the whole graph and the representations of substructures of different scales (e.g., nodes, edges, triangles).

Given a set of N training graphs \smash{$\{G_j \in \gG\}_{j=1}^{N}$} with empirical probability distribution $\mathbb{P}$ on the input space, InfoGraph behaves similarly to DGI, as it learns patch representations \smash{$\vh_j^i$} for each node $i$ in graph $G_j$ and uses them to learn a representation for the whole graph $\vs_j = \gR(\mH_j)$ through a readout function $\gR$. The network is trained by maximization of the mutual information (MI) as:
\begin{equation*}
    \theta^* = \argmax_{\theta} \sum_{G_j \in \gG} \frac{1}{|G_j|} \sum_{i \in G_j} \text{MI}(\vh_j^j, \vs_j)
\end{equation*}
The authors use the Jensen-Shannon MI estimator \citep{nowozin2016fgan}:
\begin{equation*}
    \text{MI}(\vh_j^j, \vs_j) \defeq \E_{\mathbb{P}}\left[-\sigma\left(-T(\vh_j^i,\vs_j)\right)\right] -  \E_{\mathbb{P}\times\tilde{\mathbb{P}}}\left[\sigma\left(T(\tilde{\vh}_j^i,\vs_j)\right)\right],
\end{equation*}
where $\sigma$ is the softplus function, $\vh_j^i \sim \mathbb{P}$ is an input sample, $\tilde{\vh}_j^i \sim \tilde{\mathbb{P}}=\mathbb{P}$ is a ``negative'' sample generated using all possible combinations of global and local patch representations across all graph instances in a batch, and $T$ is a discriminator.
Since $\vs_j$ is encouraged to have high MI with patches that contain information at all scales, this favors encoding aspects of the data that are shared across patches and aspects that are shared across scales. Differently from DGI \citep{velickovic2018graph}, InfoGraph uses GIN \citep{xu2018how} as graph convolutional encoders, and use the sum as the aggregator in $\gR$ instead of the mean.

\textbf{GraphCL} \citep{you2020graph} learns representation of graph data by proposing four different types of graph-level data augmentations techniques, namely node dropping, edge perturbation, attribute masking and subgraph sampling.

Given a set of N graphs \smash{$\{G_j \in \gG\}_{j=1}^{N}$}, the authors formulate the augmented graph \smash{$\hat{G} \sim q(\hat{G}|G)$}, where $q(\cdot|G)$ is the augmentation distribution conditioned on the original graph. As common in contrastive learning, the authors propose to generate two augmented views \smash{$\hat{G}_i \sim q_i(\cdot|G)$} and \smash{$\hat{G}_j \sim q_j(\cdot|G)$}. A GNN-based encoder $f$ extracts graph-level representations vectors $\vh_i, \vh_j$ for the augmented graphs \smash{$\hat{G}_i, \hat{G}_j$}, and a non-linear transformation $g$ projects the representations to another latent space where the contrastive loss is computed. The parameters of the network are optimized via minimization of the NT-Xent loss \cref{eq:L_cntr}. 
The authors claim that GraphCL boost adversarial robustness, but their evaluation is only based on synthetic dataset, hence not capturing the complexity of real-world scenarios.

In \textbf{GCA}, \citet{zhu2021graph} revisit the concept of augmentations in graph contrastive learning by arguing that data augmentation schemes should preserve intrinsic structures and attributes of graphs, which will force the model to learn representations that are insensitive to perturbation on unimportant nodes and edges. They propose a novel graph contrastive representation learning method with adaptive augmentation that incorporates various priors for topological and semantic aspects of the graph.

For \textit{topology-level augmentations}, the authors propose to corrupt the input graph by randomly removing edges. They sample a modified edge set \smash{$\tilde{E}$} from the original set of edges $E$ as \smash{$P\{(u,v) \tilde{E}\} = 1 - \rho_{uv}$}, where $(u,v) \in E$ and $\rho_{uv}$ is the probability of removing $(u,v)$. $\rho_{uv}$ should reflect the importance of edge $(u,v)$ in the graph, hence augmentations are more likely to corrupt unimportant edges while keeping important ones. The authors propose to compute $\rho_{uv}$ based on (i) \textit{degree centrality}, (ii) \textit{eigenvector centrality}, and (iii) \textit{PageRank centrality}.

Regarding \textit{node-attribute-level} augmentations, the authors perform random node attributes masking. Each node feature vector is perturbed as $\tilde{\vx} = \vx \circ \vm$, where \smash{$\vm \in \{0,1\}^d$, $\evm_i \sim \text{Bernoulli}(1 - \rho_i)$}. Similarly to $\rho_{uv}$, $\rho_i$ should capture the importance of the $i-th$ feature dimension.
To do so, they define specific weights for each feature. 
The training procedure follows the same pipeline as the one introduced by \citep{you2020graph}, where the types of augmentations are the only difference.

In \textbf{AD-GCL}, \citet{suresh2021adversarial} argue that related works in GCL are based on pre-determined data augmentation strategies that may capture redundand information about the graph. The author propose to optimize adversarial graph augmentation strategies used in GCL by minimization of the information bottleneck \citep{tishby2000information, tishby2015deep}.

Given a graph \smash{$G \in \gG$, $T(G)$} denotes a graph data augmentation of \smash{$G$}, which is a distribution defined over \smash{$\gG$} and conditioned on \smash{$G$}. \smash{$t(G) \in \gG$} is a sample of \smash{$T(G)$}. Let \smash{$\gT$} denote a family of different graph data augmentations \smash{$T_\phi(\cdot)$}, with \smash{$T_\phi(\cdot)$} being a specific augmentation scheme with parameters \smash{$\phi$}.
AD-GCL optimizes the following objective over a graph data augmentation family \smash{$\gT$}:
\begin{equation*}
    \min_{T \in \gT} \max_{f} \text{MI}\left( f(G); f(t(G)) \right), \quad \text{where } G \sim \mathbb{P}_\gG, t(G) \sim T(G),
\end{equation*}
where $f$ is a graph neural network encoder. Compared to the two graph augmentations usually adopted in GCL, AD-GCL views the original graph $G$ as the anchor while pushing its perturbation $T(G)$ as far from the anchor as it can. The automatic search over \smash{$T \in \gT$} saves a great deal of effort evaluating different combinations of graph augmentations.

For each graph \smash{$G=(V,E)$}, the sample \smash{$t(G) \sim T_\phi(\cdot)$} is a graph that shares the same node set with $G$ while the edge set of \smash{$t(G)$} is only a subset of $E$. Each edge \smash{$e \in E$} is associated with a random variable \smash{$p_e \sim \text{Bernoulli}(\omega_e)$}, where \smash{$e \in t(G)$} if \smash{$p_e = 1$}, and is dropped otherwise. To train $T(G)$ in an end-to-end fashion, the Bernoulli weights \smash{$\omega_e$} are parameterized through a GNN \textit{augmenter}. and the discrete $p_e$ are relaxed to be continuous variables in $[0, 1]$. Gumbel-Max reparametrization trick \citep{maddison2017concrete, jang2017categorical} is use to perform sampling.

    \section{Adversarial attacks}
\label{app:attacks}

In our evaluation, we consider various \textit{adaptive} gradient-based adversarial attack schemes, such as PGD \citep{xu2019topology}, PR-BCD and GR-BCD \citep{geisler2021robustness}, and \textit{static} adversarial attacks as simple baselines. While we only consider a single global budget \(\Delta\), it is straightforward to include more sophisticated constraints when beneficial for the application at hand \citep{gosch2023revisiting, gosch2023adversarial}.

\textbf{Random Edge Flipping} is a straightforward attack scheme in which edges are randomly flipped with a given perturbation probability $\rho$. Formally, this can be expressed as \smash{$\tilde{\emA}_{ij} = \text{flip}(\emA_{ij}, \rho)$}, where $\emA_{ij}$ represents the edge $(i,j)$ in the original adjacency matrix of the graph, and \smash{$\tilde{\emA}_{ij}$} is the perturbed edge.

\textbf{Projected Gradient Descent (PGD)} \citep{xu2019topology} uses continuous relaxation to approximate \cref{eq:adv_atk}. Specifically, the adjacency is relaxed from \smash{\(\{0,1\}^{n \times n} \to [0,1]^{n \times n}\)}. Due to this relaxation, first-order methods can be applied as long as the targeted model can handle weighted edges (i.e., \smash{\(g_{\phi^*}(f_{\theta^*}(\mX, \tilde{\mA}))\))}. Following, gradient descent is used with an additional projection to ensure that the budget is not exceeded and that the value range \([0,1]\) is not violated. The last step of the attack, then, discretizes the perturbed adjacency \smash{\([0,1]^{n \times n} \to \{0,1\}^{n \times n}\)}.

\textbf{Projected Randomized Block Coordinate Descent (PR-BCD)} \citep{geisler2021robustness} works similar to the PGD attack of \citet{xu2019topology}, while avoiding its quadratic space complexity. The quadratic memory requirement arises from the fact that \(\mA\) has up to \(n^2\) non-zero entries and PGD (typically) optimizes over all of them. To circumvent the quadratic cost, PR-BCD performs the gradient update only for a random and non-contiguous block of entries in \(\mA\) at a time. Thereafter, in a survival-of-the-fittest manner, the random block is resampled. Specifically, relevant entries are kept in the next block, and the remainder is discarded as well as resampled. This way, the additional space complexity is linear in the random block size. In other words, PR-BCD uses randomization in the gradient update to obtain scalability. \citet{geisler2021robustness} apply the PR-BCD attack to graphs of up to 100 million nodes.

\textbf{Greedy Randomized Block Coordinate Descent (GR-BCD)} \citep{geisler2021robustness} works similar to PR-BCD, with the notable exception of pursuing a greedy objective. First, the budget \(\Delta\) is distributed over the desired number of greedy updates. Then, in each greedy update, the desired amount of edges in the randomly drawn block is flipped. The edges to be flipped are chosen based on the gradient \smash{\(\nabla \gL_\text{atk}(g_{\phi^*}(f_{\theta^*}(\mX, \tilde{\mA})))\)}. Due to the greediness, GR-BCD does not require the random block size to be larger than \(\Delta\), and, thus, GR-BCD is even more scalable than PR-BCD.

    \section{Datasets}
For graph classification, we utilize three datasets of graphs: PROTEINS, NCI1, and DD \citep{morris2020tudataset}. For node classification, we conduct experiments on well-established node classification datasets, namely Cora, Citeseer \cite{sen2008collective}, Pubmed \citep{namata2012querydriven}. We also include in our analysis the \textit{large-scale} graph OGB-arXiv \citep{hu2020open}. 

\cref{tab:datasets_gc,tab:datasets_nc} report the statistics for the datasets used in our empirical evaluation, for graph and node classification, respectively.

\begin{table}[t]
\centering
\caption{Graph classification datasets.}
\label{tab:datasets_gc}
\begin{tabular}{@{}lccc@{}}
\toprule
\textbf{Dataset} & \textbf{\# Graphs} & \textbf{\# Nodes} & \textbf{\# Edges} \\ 
& & \small{(min, max, median)} & \small{(min, max, median)} \\ \midrule
DD       & 1178  & (30, 5748, 241) & (126, 28534, 1220) \\
NCI1     & 3327  & (3, 111, 27)    & (4, 238, 58)       \\
PROTEINS & 19717 & (4, 620, 26)    & (10, 2098, 98)     \\ \bottomrule
\end{tabular}
\end{table}
\begin{table}[t]
\centering
\caption{Node classification datasets.}
\label{tab:datasets_nc}
\begin{tabular}{@{}lcccc@{}}
\toprule
\textbf{Dataset} & \textbf{\# Nodes} & \textbf{\# Edges} & \textbf{\# Features} & \textbf{\# Classes} \\ \midrule
Planetoid-Cora     & 2708   & 10556   & 1433 & 7  \\
Planetoid-Citeseer & 3327   & 9104    & 3703 & 6  \\
Planetoid-Pubmed   & 19717  & 88648   & 500  & 3  \\
OGB-arXiv          & 169343 & 1166243 & 128  & 40 \\ \bottomrule
\end{tabular}
\end{table}
    \clearpage
\section{Complete results}
\label{app:results}
Here we report the extensions of \cref{tab:res_gc,tab:res_nc} to different attack schemes. As a general observation, we can notice that \textit{adaptive} adversarial attacks (PGD, R-BCD), increase the drop in accuracies of the models in every dataset, compared to static attacks (random edge flipping).

\begin{table}[htbp]
\centering
\caption{\textbf{Graph classification} performance: clean and perturbed accuracy mean $\pm$ std., and $\relDrop$ (percentage), averaged over 15 runs with different weight initializations. Results include accuracy for different contrastive and non-contrastive models under various attack schemes and datasets. The attacks are allowed (\(\Delta\)) to change 5\% of edges. Models achieving the three highest relative drop are highlighted as \textcolor{first}{\textbf{first}}, \textcolor{second}{\textbf{second}}, and \textcolor{third}{\textbf{third}}.}
\label{tab:res_gc_complete}
\resizebox{\linewidth}{!}{%
\begin{tabular}{@{}llcccccc@{}}
\toprule
\textbf{Model} & \textbf{Attack} & \multicolumn{2}{c}{\textsc{PROTEINS}} & \multicolumn{2}{c}{\textsc{NCI1}} & \multicolumn{2}{c}{\textsc{DD}} \\ \midrule
GCN
& \textsc{Clean} & 
\scriptsize{$73.98 \pm 1.00$} & &
\scriptsize{$74.67 \pm 0.80$} & & 
\scriptsize{$70.02 \pm 0.54$} & \\
& \textsc{Rand. Edge Flip.} & 
\scriptsize{$73.41 \pm 0.97$} & $(\downarrow ~~0.78)$ &  
\scriptsize{$68.12 \pm 1.29$} & $(\downarrow ~~8.77)$ &  
\scriptsize{$60.29 \pm 2.23$} & $(\downarrow 13.98)$ \\
& \textsc{PGD} & 
\scriptsize{$68.68 \pm 0.86$} & $(\downarrow ~~7.16)$ &  
\scriptsize{$52.69 \pm 1.05$} & $(\downarrow 29.43)$ &  
\scriptsize{$51.21 \pm 2.75$} & $(\downarrow 26.85)$ \\
& \textsc{PR-BCD} & 
\scriptsize{$68.04 \pm 0.90$} & $(\downarrow ~~8.04)$ &  
\scriptsize{$33.91 \pm 1.36$} & $(\downarrow 54.59)$ &  
\scriptsize{$~~8.58 \pm 7.35$} & $(\downarrow 87.57)$ \\
& \textsc{GR-BCD} & 
\scriptsize{$72.50 \pm 1.47$} & $(\downarrow ~~2.00)$ & 
\scriptsize{$49.47 \pm 4.21$} & $(\downarrow 33.75)$ & 
\scriptsize{$57.33 \pm 2.56$} & $(\downarrow 18.24)$ \\ \cmidrule(l){2-8}
& \textsc{Min} & 
\scriptsize{$68.04 \pm 0.90$} & $(\downarrow ~~8.04)$ & 
\scriptsize{$33.91 \pm 1.36$} & $(\downarrow 54.59)$ &  
\scriptsize{$~~8.58 \pm 7.35$} & $\textcolor{first}{\mathbf{(\downarrow 87.57)}}$ \\
\midrule
GIN
& \textsc{Clean} & 
\scriptsize{$66.02 \pm 4.60$} & & 
\scriptsize{$76.04 \pm 0.91$} & &
\scriptsize{$63.44 \pm 3.19$} \\  
& \textsc{Rand. Edge Flip.} & 
\scriptsize{$59.71 \pm 4.63$} & $(\downarrow ~~9.55)$ & 
\scriptsize{$54.47 \pm 1.56$} & $(\downarrow 28.37)$ &  
\scriptsize{$56.73 \pm 3.61$} & $(\downarrow 10.57)$ \\ 
& \textsc{PGD} & 
\scriptsize{$61.12 \pm 4.76$} & $(\downarrow ~~7.42)$ &  
\scriptsize{$71.59 \pm 0.83$} & $(\downarrow ~~5.86)$ & 
\scriptsize{$58.23 \pm 3.23$} & $(\downarrow ~~8.21)$ \\
& \textsc{PR-BCD} & 
\scriptsize{$48.34 \pm 6.74$} & $(\downarrow 26.77)$ &  
\scriptsize{$38.65 \pm 2.48$} & $(\downarrow 49.17)$ &  
\scriptsize{$16.17 \pm 6.16$} & $(\downarrow 74.51)$ \\
& \textsc{GR-BCD} & 
\scriptsize{$46.05 \pm 6.86$} & $(\downarrow 30.24)$ & 
\scriptsize{$43.99 \pm 1.77$} & $(\downarrow 42.15)$ & 
\scriptsize{$34.83 \pm 8.31$} & $(\downarrow 45.10)$ \\ \cmidrule(l){2-8}
& \textsc{Min} & 
\scriptsize{$46.05 \pm 6.86$} & $(\downarrow 30.24)$ & 
\scriptsize{$38.65 \pm 2.48$} & $(\downarrow 49.17)$ &   
\scriptsize{$16.17 \pm 6.16$} & $\textcolor{third}{\mathbf{(\downarrow 74.51)}}$ \\
\midrule
InfoGraph
& \textsc{Clean} & 
\scriptsize{$63.93 \pm 6.43$} & & 
\scriptsize{$66.87 \pm 3.72$} & & 
\scriptsize{$64.77 \pm 2.41$} \\
& \textsc{Rand. Edge Flip.} &  
\scriptsize{$56.85 \pm 6.22$} & $(\downarrow 11.10)$ &  
\scriptsize{$50.88 \pm 3.37$} & $(\downarrow 23.83)$ &  
\scriptsize{$58.84 \pm 2.66$} & $(\downarrow ~~9.15)$ \\
& \textsc{PGD} & 
\scriptsize{$51.65 \pm 7.50$} & $(\downarrow 19.42)$ &  
\scriptsize{$30.68 \pm 4.89$} & $(\downarrow 54.18)$ &  
\scriptsize{$55.12 \pm 3.81$} & $(\downarrow 14.87)$ \\
& \textsc{PR-BCD} & 
\scriptsize{$31.53 \pm 7.67$} & $(\downarrow 50.83)$ &  
\scriptsize{$14.71 \pm 4.13$} & $(\downarrow 78.10)$ &  
\scriptsize{$20.11 \pm 4.24$} & $(\downarrow 68.92)$ \\
& \textsc{GR-BCD} & 
\scriptsize{$47.23 \pm 7.55$} & $(\downarrow 26.19)$ & 
\scriptsize{$41.13 \pm 4.40$} & $(\downarrow 38.45)$ & 
\scriptsize{$24.99 \pm 5.33$} & $(\downarrow 61.44)$ \\ \cmidrule(l){2-8}
& \textsc{Min} & 
\scriptsize{$31.53 \pm 7.67$} & $\textcolor{second}{\mathbf{(\downarrow 50.83)}}$ &  
\scriptsize{$14.71 \pm 4.13$} & $\textcolor{second}{\mathbf{(\downarrow 78.10)}}$ &  
\scriptsize{$20.11 \pm 4.24$} & $(\downarrow 68.92)$ \\
\midrule
GraphCL
& \textsc{Clean} & 
\scriptsize{$65.93 \pm 4.02$} & & 
\scriptsize{$73.09 \pm 3.50$} & & 
\scriptsize{$68.85 \pm 4.83$} \\
& \textsc{Rand. Edge Flip.} &  
\scriptsize{$61.62 \pm 4.94$} & $(\downarrow ~~6.53)$ &  
\scriptsize{$59.43 \pm 2.36$} & $(\downarrow 18.69)$ &  
\scriptsize{$58.43 \pm 5.11$} & $(\downarrow ~~8.48)$ \\ 
& \textsc{PGD} & 
\scriptsize{$60.86 \pm 4.02$} & $(\downarrow ~~7.68)$ &  
\scriptsize{$68.32 \pm 2.54$} & $(\downarrow 10.63)$ &  
\scriptsize{$58.35 \pm 6.21$} & $(\downarrow ~~8.61)$ \\
& \textsc{PR-BCD} & 
\scriptsize{$49.48 \pm 6.39$} & $(\downarrow 24.95)$ &  
\scriptsize{$32.20 \pm 3.48$} & $(\downarrow 55.94)$ &  
\scriptsize{$15.74 \pm 7.66$} & $(\downarrow 75.34)$ \\
& \textsc{GR-BCD} & 
\scriptsize{$40.29 \pm 9.61$} & $(\downarrow 38.89)$ & 
\scriptsize{$37.09 \pm 3.37$} & $(\downarrow 49.26)$ & 
\scriptsize{$32.82 \pm 6.93$} & $(\downarrow 48.60)$ \\ \cmidrule(l){2-8}
& \textsc{Min} & 
\scriptsize{$40.29 \pm 9.61$} & $\textcolor{third}{\mathbf{(\downarrow 38.89)}}$ & 
\scriptsize{$32.20 \pm 3.48$} & $\textcolor{third}{\mathbf{(\downarrow 55.94)}}$ &  
\scriptsize{$15.74 \pm 7.66$} & $\textcolor{second}{\mathbf{(\downarrow 75.34)}}$ \\
\midrule
AD-GCL
& \textsc{Clean} & 
\scriptsize{$73.66 \pm 1.36$} & & 
\scriptsize{$71.79 \pm 0.94$} & & 
\scriptsize{$76.71 \pm 2.07$} \\
& \textsc{Rand. Edge Flip.} &  
\scriptsize{$65.37 \pm 2.24$} & $(\downarrow 11.25)$ &  
\scriptsize{$58.12 \pm 1.49$} & $(\downarrow 19.04)$ &  
\scriptsize{$75.18 \pm 2.03$} & $(\downarrow ~~1.95)$ \\
& \textsc{PGD} & 
\scriptsize{$61.44 \pm 2.57$} & $(\downarrow 16.59)$ &  
\scriptsize{$50.32 \pm 2.32$} & $(\downarrow 29.91)$ &  
\scriptsize{$44.36 \pm 5.41$} & $(\downarrow 42.18)$ \\
& \textsc{PR-BCD} & 
\scriptsize{$26.31 \pm 4.33$} & $(\downarrow 64.28)$ & 
\scriptsize{$14.35 \pm 1.34$} & $(\downarrow 80.01)$ &  
\scriptsize{$29.17 \pm 5.27$} & $(\downarrow 61.71)$ \\
& \textsc{GR-BCD} & 
\scriptsize{$63.96 \pm 2.90$} & $(\downarrow 13.17)$ & 
\scriptsize{$54.52 \pm 1.44$} & $(\downarrow 24.06)$ & 
\scriptsize{$37.48 \pm 7.03$} & $(\downarrow 50.98)$ \\ \cmidrule(l){2-8}
& \textsc{Min} & 
\scriptsize{$26.31 \pm 4.33$} & $\textcolor{first}{\mathbf{(\downarrow 64.28)}}$ & 
\scriptsize{$14.35 \pm 1.34$} & $\textcolor{first}{\mathbf{(\downarrow 80.01)}}$ &  
\scriptsize{$29.17 \pm 5.27$} & $(\downarrow 61.71)$ \\
\bottomrule
\end{tabular}
}
\end{table}
\begin{table}[htbp]
\centering
\caption{\textbf{Node classification} performance: clean and perturbed accuracy mean $\pm$ std., and $\relDrop$ (percentage), averaged over 15 runs with different weight initializations. Results include accuracy for different contrastive and non-contrastive models under various attack schemes and datasets. The attacks are allowed (\(\Delta\)) to change 5\% of edges. Models achieving the three highest relative drop are highlighted as \textcolor{first}{\textbf{first}}, \textcolor{second}{\textbf{second}}, and \textcolor{third}{\textbf{third}}.}
\label{tab:res_nc_complete}
\resizebox{\linewidth}{!}{%
\begin{tabular}{@{}llcccccccc@{}}
\toprule
\textbf{Model} & \textbf{Attack} & \multicolumn{2}{c}{\textsc{Cora}} & \multicolumn{2}{c}{\textsc{Citeseer}} & \multicolumn{2}{c}{\textsc{Pubmed}} & \multicolumn{2}{c}{\textsc{OGB-arXiv}} \\ \midrule
GCN 
& \textsc{Clean} & 
\scriptsize{$77.57 \pm 1.11$} & &
\scriptsize{$63.99 \pm 1.27$} & &
\scriptsize{$75.31 \pm 0.80$} & &
\scriptsize{$68.22 \pm 1.15$} \\
& \textsc{Rand. Edge Flip.} &
\scriptsize{$76.55 \pm 1.19$} & $(\downarrow ~~1.32)$ &
\scriptsize{$62.65 \pm 1.58$} & $(\downarrow ~~2.09)$ & 
\scriptsize{$74.34 \pm 0.88$} & $(\downarrow ~~1.31)$ & 
\scriptsize{$66.07 \pm 1.09$} & $(\downarrow ~~3.15)$ \\
& \textsc{PR-BCD} & 
\scriptsize{$59.35 \pm 1.70$} & $(\downarrow 23.50)$ & 
\scriptsize{$47.69 \pm 2.06$} & $(\downarrow 25.46)$ & 
\scriptsize{$57.01 \pm 1.28$} & $(\downarrow 24.31)$ & 
\scriptsize{$52.54 \pm 1.08$} & $(\downarrow 22.99)$ \\
& \textsc{GR-BCD} & 
\scriptsize{$70.37 \pm 1.61$} & $(\downarrow ~~9.29)$ &
\scriptsize{$55.31 \pm 2.21$} & $(\downarrow 13.57)$ &
\scriptsize{$64.33 \pm 1.97$} & $(\downarrow 14.59)$ &
\scriptsize{$49.58 \pm 1.37$} & $(\downarrow 27.33)$ \\ \cmidrule(l){2-10}
& \textsc{Min} & 
\scriptsize{$59.35 \pm 1.70$} & $\textcolor{second}{\mathbf{(\downarrow 23.50)}}$ & 
\scriptsize{$47.69 \pm 2.06$} & $\textcolor{second}{\mathbf{(\downarrow 25.46)}}$ & 
\scriptsize{$57.01 \pm 1.28$} & $\textcolor{third}{\mathbf{(\downarrow 24.31)}}$ &
\scriptsize{$49.58 \pm 1.37$} & $\textcolor{second}{\mathbf{(\downarrow 27.33)}}$ \\
\midrule
DGI 
& \textsc{Clean} & 
\scriptsize{$83.24 \pm 1.37$} & &
\scriptsize{$72.91 \pm 1.99$} & &
\scriptsize{$81.46 \pm 0.79$} & &
\scriptsize{$60.18 \pm 0.72$} \\
& \textsc{Rand. Edge Flip.} &
\scriptsize{$82.65 \pm 1.25$} & $(\downarrow ~~0.71)$ & 
\scriptsize{$72.64 \pm 1.96$} & $(\downarrow ~~0.39)$ & 
\scriptsize{$80.72 \pm 0.844$} & $(\downarrow ~~0.91)$ & 
\scriptsize{$59.14 \pm 0.76$} & $(\downarrow 1.72)$ \\
& \textsc{PR-BCD} & 
\scriptsize{$75.69 \pm 1.86$} & $(\downarrow ~~9.08)$ & 
\scriptsize{$67.47 \pm 2.13$} & $(\downarrow ~~7.47)$ & 
\scriptsize{$77.19 \pm 0.92$} & $(\downarrow ~~5.23)$ & 
\scriptsize{$53.28 \pm 0.69$} & $(\downarrow 11.45)$ \\
& \textsc{GR-BCD} & 
\scriptsize{$80.13 \pm 1.55$} & $(\downarrow ~~3.73)$ &
\scriptsize{$70.36 \pm 2.44$} & $(\downarrow ~~3.52)$ &
\scriptsize{$77.12 \pm 0.89$} & $(\downarrow ~~5.33)$ &
\scriptsize{$54.19 \pm 0.87$} & $(\downarrow ~~9.96)$ \\ \cmidrule(l){2-10}
& \textsc{Min} & 
\scriptsize{$75.69 \pm 1.86$} & $(\downarrow ~~9.08)$ & 
\scriptsize{$67.47 \pm 2.13$} & $(\downarrow ~~7.47)$ & 
\scriptsize{$77.12 \pm 0.89$} & $(\downarrow ~~5.33)$ &
\scriptsize{$53.28 \pm 0.69$} & $\textcolor{third}{\mathbf{(\downarrow 11.45)}}$ \\
\midrule
GraphCL 
& \textsc{Clean} & 
\scriptsize{$71.99 \pm 1.35$} & &
\scriptsize{$59.57 \pm 1.45$} & &
\scriptsize{$74.29 \pm 1.75$} & &
\scriptsize{$62.78 \pm 0.44$} \\
& \textsc{Rand. Edge Flip.} &
\scriptsize{$70.86 \pm 1.27$} & $(\downarrow ~~1.57)$ & 
\scriptsize{$58.61 \pm 1.76$} & $(\downarrow ~~1.62)$ & 
\scriptsize{$72.86 \pm 1.52$} & $(\downarrow ~~1.92)$ & 
\scriptsize{$60.37 \pm 0.43$} & $(\downarrow ~~3.83)$ \\
& \textsc{PR-BCD} & 
\scriptsize{$54.71 \pm 1.46$} & $(\downarrow 24.00)$ & 
\scriptsize{$46.53 \pm 1.87$} & $(\downarrow 21.89)$ & 
\scriptsize{$58.71 \pm 1.45$} & $(\downarrow 20.96)$ & 
\scriptsize{$49.07 \pm 0.37$} & $(\downarrow 21.84)$ \\
& \textsc{GR-BCD} & 
\scriptsize{$60.15 \pm 1.26$} & $(\downarrow 16.44)$ &
\scriptsize{$48.99 \pm 1.75$} & $(\downarrow 17.77)$ &
\scriptsize{$53.99 \pm 1.77$} & $(\downarrow 27.33)$ &
\scriptsize{$36.57 \pm 0.55$} & $(\downarrow 41.75)$ \\ \cmidrule(l){2-10}
& \textsc{Min} & 
\scriptsize{$54.71 \pm 1.46$} & $\textcolor{first}{\mathbf{(\downarrow 24.00)}}$ & 
\scriptsize{$46.53 \pm 1.87$} & $\textcolor{third}{\mathbf{(\downarrow 21.89)}}$ & 
\scriptsize{$53.99 \pm 1.77$} & $\textcolor{second}{\mathbf{(\downarrow 27.33)}}$ &
\scriptsize{$36.57 \pm 0.55$} & $\textcolor{first}{\mathbf{(\downarrow 41.75)}}$ \\
\midrule
GCA 
& \textsc{Clean} & 
\scriptsize{$79.07 \pm 1.36$} & &
\scriptsize{$60.14 \pm 2.24$} & &
\scriptsize{$78.62 \pm 0.98$} & &
\multicolumn{2}{c}{\multirow{5}{*}{OOM}} \\
& \textsc{Rand. Edge Flip.} &
\scriptsize{$78.30 \pm 1.63$} & $(\downarrow ~~0.98)$ & 
\scriptsize{$59.21 \pm 2.15$} & $(\downarrow ~~1.54)$ & 
\scriptsize{$76.09 \pm 1.07$} & $(\downarrow ~~3.22)$ & \\
& \textsc{PR-BCD} & 
\scriptsize{$61.63 \pm 3.18$} & $(\downarrow 22.08)$ & 
\scriptsize{$43.61 \pm 2.13$} & $(\downarrow 27.51)$ & 
\scriptsize{$56.77 \pm 1.69$} & $(\downarrow 27.80)$ & \\
& \textsc{GR-BCD} & 
\scriptsize{$72.59 \pm 2.75$} & $(\downarrow ~~8.22)$ &
\scriptsize{$50.45 \pm 2.65$} & $(\downarrow 16.14)$ &
\scriptsize{$57.67 \pm 1.87$} & $(\downarrow 26.67)$ & \\ \cmidrule(l){2-8}
& \textsc{Min} & 
\scriptsize{$61.63 \pm 3.18$} & $\textcolor{third}{\mathbf{(\downarrow 22.08)}}$ & 
\scriptsize{$43.61 \pm 2.13$} & $\textcolor{first}{\mathbf{(\downarrow 27.51)}}$ & 
\scriptsize{$56.77 \pm 1.69$} & $\textcolor{first}{\mathbf{(\downarrow 27.80)}}$ & \\
\bottomrule
\end{tabular}
}
\end{table}
    \clearpage
\section{Hyperparameters}
\label{app:hparams}
We report the hyperparameters we used for our evaluation in \cref{tab:hparams}.
\begin{table}[htbp]
\centering
\caption{Models hyperparameters.}
\label{tab:hparams}
\begin{tabular}{@{}llcccccc@{}}
\toprule
\textbf{Model} & \textbf{Dataset} & \multicolumn{6}{c}{\textbf{Hyperparameters}} \\ 
               &                  & lr & epochs & patience & dropout & layers & hid. dim. \\ \midrule
GCN
& \textsc{Cora}      & 1e-2 & 200 & 10 & 0.5 & 2 & 16 \\
& \textsc{Citeseer}  & 1e-2 & 200 & 10 & 0.5 & 2 & 16 \\
& \textsc{Pubmed}    & 1e-2 & 200 & 10 & 0.5 & 2 & 16 \\
& \textsc{OGB-arXiv} & 1e-2 & 500 & 10 & 0.5 & 3 & 256 \\
& \textsc{PROTEINS}  & 5e-3 & 50 & & & 4 & 128 \\
& \textsc{NCI1}      & 5e-3 & 50 & & & 4 & 128 \\
& \textsc{DD}        & 5e-3 & 50 & & & 4 & 32 \\
\midrule 
GIN
& \textsc{PROTEINS}  & 1e-3 & 10 & & & 8 & 512 \\
& \textsc{NCI1}      & 1e-4 & 10 & & & 12 & 512 \\
& \textsc{DD}        & 1e-4 & 20 & & & 4 & 32 \\
\midrule
DGI
& \textsc{Cora}      & 1e-3 & 1000 & 20 & & 1 & 512\\
& \textsc{Citeseer}  & 1e-3 & 1000 & 20 & & 1 & 512 \\
& \textsc{Pubmed}    & 1e-3 & 1000 & 20 & & 1 & 256 \\
& \textsc{OGB-arXiv} & 1e-3 & 1000 & 20 & & 2 & 512 \\
\midrule
GraphCL
& \textsc{Cora}      & 1e-3 & 1000 & 20 & & 1 & 512\\
& \textsc{Citeseer}  & 1e-3 & 1000 & 20 & & 1 & 512\\
& \textsc{Pubmed}    & 1e-3 & 1000 & 20 & & 1 & 512\\
& \textsc{OGB-arXiv} & 1e-3 & 1000 & 20 & & 1 & 512 \\
& \textsc{PROTEINS}  & 1e-3 & 10 & & & 8 & 512 \\
& \textsc{NCI1}      & 1e-4 & 10 & & & 12 & 512 \\
& \textsc{DD}        & 1e-4 & 20 & & & 4 & 32 \\
\midrule
GCA
& \textsc{Cora}      & 1e-3 & 500 & 20 & & & 256 \\
& \textsc{Citeseer}  & 1e-3 & 500 & 20 & & & 256 \\
& \textsc{Pubmed}    & 1e-3 & 500 & 20 & & & 256 \\
& \textsc{OGB-arXiv} & 1e-3 & 500 & 20 & & & 256 (OOM) \\
\midrule 
InfoGraph
& \textsc{PROTEINS}  & 1e-3 & 100 & & & 8 & 256 \\
& \textsc{NCI1}      & 1e-3 & 100 & & & 8 & 256 \\
& \textsc{DD}        & 1e-3 & 100 & & & 8 & 256 \\
\midrule 
AD-GCL
& \textsc{PROTEINS}  & 1e-2 & 150 & 20 & 0.5 & 5 & 32 \\
& \textsc{NCI1}      & 1e-2 & 150 & 20 & 0.5 & 5 & 32 \\
& \textsc{DD}        & 1e-2 & 150 & 20 & 0.5 & 5 & 32 \\
\bottomrule
\end{tabular}
\end{table}
\end{document}